\pdfoutput=1
\documentclass[11pt]{article}

\usepackage[]{StyleFiles/EMNLP2023}

\usepackage{times}
\usepackage{latexsym}

\usepackage{amsmath}
\usepackage{caption}
\usepackage{subcaption}
\usepackage{hyperref}
\usepackage{url}
\usepackage{lipsum}
\usepackage{graphicx}
\usepackage{tabularx}
\usepackage{array}
\usepackage{multirow}
\usepackage{tikz}
\usepackage{pgf-pie}

\usepackage{graphicx}
\usepackage{tabularx}
\usepackage{soul}
\usepackage{caption}
\usepackage{subcaption}
\usepackage{latexsym}
\usepackage{pgfplots}
\usepackage{amsmath}
\usepackage{caption}
\usepackage{subcaption}
\usepackage{hyperref}
\usepackage{url}
\usepackage{lipsum}
\usepackage{graphicx}
\usepackage{tabularx}
\usepackage{array}
\usepackage{multirow}

\usepackage[T1]{fontenc}
\usepackage[utf8]{inputenc}

\usepackage{microtype}

\usepackage{inconsolata}

\title{The Battle of LLMs: A Comparative Study in Conversational QA Tasks}
\author{\textbf{Aryan Rangapur} \quad \textbf{Aman Rangapur} \\
        Vellore Institute of Technology AP, India \\
        Illinois Institute of Technology, Chicago, IL, USA \\
        \texttt{aryan.22bce7399@vitapstudent.ac.in, arangapur@hawk.iit.edu}
        }
\begin{document}
\maketitle
\begin{abstract}

Large language models have gained considerable interest for their impressive performance on various tasks. Within this domain, ChatGPT and GPT-4, developed by OpenAI, and the Gemini, developed by Google, have emerged as particularly popular among early adopters. Additionally, Mixtral by Mistral AI and Claude by Anthropic are newly released, further expanding the landscape of advanced language models. These models are viewed as disruptive technologies with applications spanning customer service, education, healthcare, and finance. More recently, Mistral has entered the scene, captivating users with its unique ability to generate creative content. Understanding the perspectives of these users is crucial, as they can offer valuable insights into the potential strengths, weaknesses, and overall success or failure of these technologies in various domains. This research delves into the responses generated by ChatGPT, GPT-4, Gemini, Mixtral and Claude across different Conversational QA corpora. Evaluation scores were meticulously computed and subsequently compared to ascertain the overall performance of these models. Our study pinpointed instances where these models provided inaccurate answers to questions, offering insights into potential areas where they might be susceptible to errors. In essence, this research provides a comprehensive comparison and evaluation of these state-of-the-art language models, shedding light on their capabilities while also highlighting potential areas for improvement.
\end{abstract}

\section{Introduction}
The advent of large language models has routed a new era in the field of artificial intelligence. These sophisticated neural network models are capable of generating text that mirrors specific tones and content. Trained on extensive datasets, they predict the most appropriate continuation for a given prompt, thereby producing outputs that are remarkably human-like \cite{https://doi.org/10.48550/arxiv.2212.05856}.

The models that have emerged as leaders in this space are ChatGPT, developed by OpenAI \cite{radford2018improving}, Gemini, developed by Google \cite{team2023gemini}, Mixtral developed by Mistral AI team \cite{jiang2023mistral} and Claude developed by Anthropic. All of these models have gained immense popularity due to their impressive performance on a variety of language tasks. They are large pre-trained language models that employ deep learning techniques to generate responses to natural language queries. Their ability to understand and generate coherent responses has made them invaluable tools in a wide range of applications, including chatbots, language translation, question-answering systems, and creative content generation \cite{https://doi.org/10.48550/arxiv.2303.01194}.

These models differ from conventional chatbots in several ways. They have the ability to recall previous conversations with users, decline unsuitable requests, and correct inaccurate responses. They can provide detailed answers, suggestions, and explanations to complex queries, such as coding, optimization, and layout issues \cite{https://doi.org/10.48550/arxiv.2303.01194}. Owing to their superior capabilities, these models have garnered significant attention and user base, surpassing other well-known online platforms. They have been pre-trained on a massive corpus of text data and have shown a remarkable ability to generate human-like text in response to natural language inputs \cite{https://doi.org/10.48550/arxiv.2303.01248}.

The pre-training process of these models involves several stages: unsupervised pre-training, supervised fine-tuning, and having a “human-in-the-loop” to fine tune the model’s ability to understand human instruction better. In the unsupervised pre-training stage, the models are trained on a massive dataset of text to learn the patterns and structure of natural language.Subsequently, in the supervised fine-tuning stage, the models undergo further training on labeled datasets, refining their understanding and adapting to specific tasks. The "human-in-the-loop" phase follows, incorporating human feedback to enhance the model's ability to comprehend and respond effectively to nuanced instructions. This iterative refinement ensures improved performance and alignment with human communication nuances \cite{https://doi.org/10.48550/arxiv.2302.14600}.

These Large Language Models are groundbreaking technologies that have the potential to transform the way we interact with machines. They can be used for a variety of applications, including chatbots, language translation, text summarization, and creative content generation \cite{https://doi.org/10.48550/arxiv.2303.01157}. Numerous industries have already adopted these technologies, including e-commerce, customer service, and healthcare, to provide personalized and efficient customer support.

This research paper aims to explore the performance of ChatGPT, Gemini, Mixtral and Claude and their potential use in various domains. The study analyzes the accuracy and consistency of the model’s responses to different datasets and investigates the areas where the models may be prone to error. We aim to analyze the reliability of the model’s output for conversational QA tasks \cite{https://doi.org/10.48550/arxiv.1808.07042,https://doi.org/10.48550/arxiv.2110.08222,https://doi.org/10.48550/arxiv.2107.02153,https://doi.org/10.48550/arxiv.1904.04365}. To achieve this, we developed a pipeline that generates large-scale responses and conducted a thorough comparison between the model’s responses and existing QA corpora. We calculated various scores like BLEU, ROUGE, etc. of the model’s responses to assess the golden ratio and fluency of their output. We also evaluated the models using the Chain of Thought method, as well as Zero Shot and 3-shot learning \cite{wei2022chain,sung2018learning}. These methods allowed us to assess the model’s ability to maintain context over a series of interrelated queries and quickly adapt to new tasks with minimal examples.
%We also explored the potential ethical implications and challenges posed by these models, providing a comprehensive overview of the current state of large language models and their impact on the field of natural language processing.

\section{Methodology}
In the course of our investigation, we meticulously devised and implemented a sophisticated pipeline aimed at harnessing the potent capabilities of ChatGPT, Gemini, Mixtral and Claude to produce expansive responses at scale. This meticulously crafted pipeline is structured around two pivotal modules, specifically the question generation module and the response generation module.

The question generation module serves the crucial function of crafting a diverse array of questions that aptly encapsulate the essence of typical conversational QA tasks. To ensure a comprehensive coverage of a myriad of topics, we judiciously employed a suite of techniques including paraphrasing, augmentation, and strategic sampling from existing QA corpora. Subsequently, these meticulously generated questions were then used to interrogate the language models, soliciting responses.

On the flip side, the response generation module harnesses the advanced language generation capabilities inherent in the language models to construct coherent and relevant responses to the questions emanating from the question generation module. The resultant responses underwent meticulous evaluation to gauge their relevance and specificity to the corresponding questions.

To validate the efficacy of our pipeline, we subjected it to rigorous evaluation using four widely recognized datasets: CoQA \cite{https://doi.org/10.48550/arxiv.1808.07042}, DialFact \cite{https://doi.org/10.48550/arxiv.2110.08222}, FaVIQ \cite{https://doi.org/10.48550/arxiv.2107.02153}, and CoDAH \cite{https://doi.org/10.48550/arxiv.1904.04365}. These datasets are widely acknowledged benchmarks for assessing conversational QA tasks, encompassing a diverse spectrum of topics and domains. Additional details of these datasets are discussed in Appendix \ref{addn_dataset}.

\begin{figure*}[!t]
    \centering
    \begin{subfigure}[t]{0.39\textwidth}
        \centering
        \includegraphics[width=\textwidth]{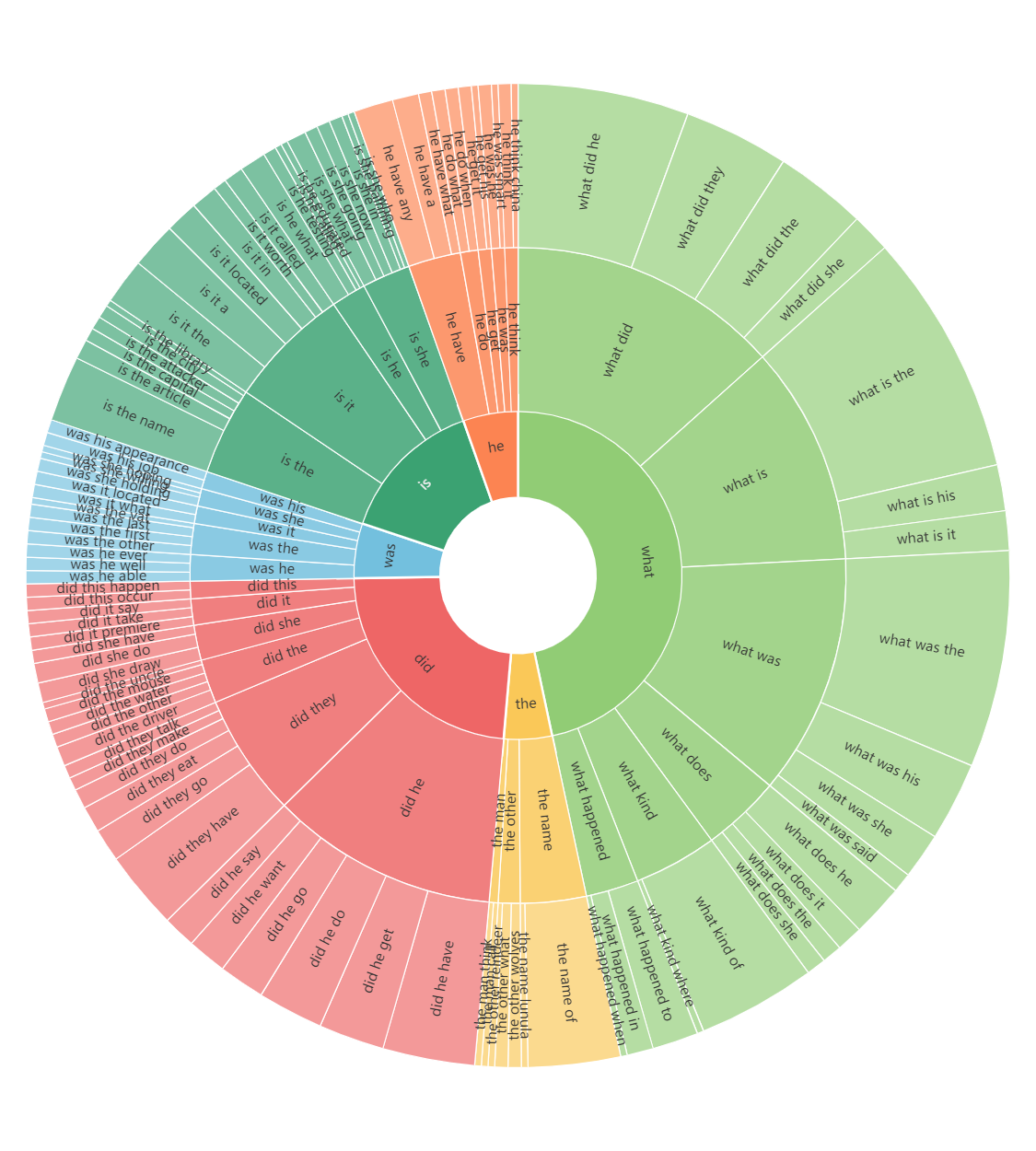}
        \caption{CoQA}
        \label{fig:sub1}
    \end{subfigure}
    \hfill
    \begin{subfigure}[t]{0.39\textwidth}
        \centering
        \includegraphics[width=\textwidth]{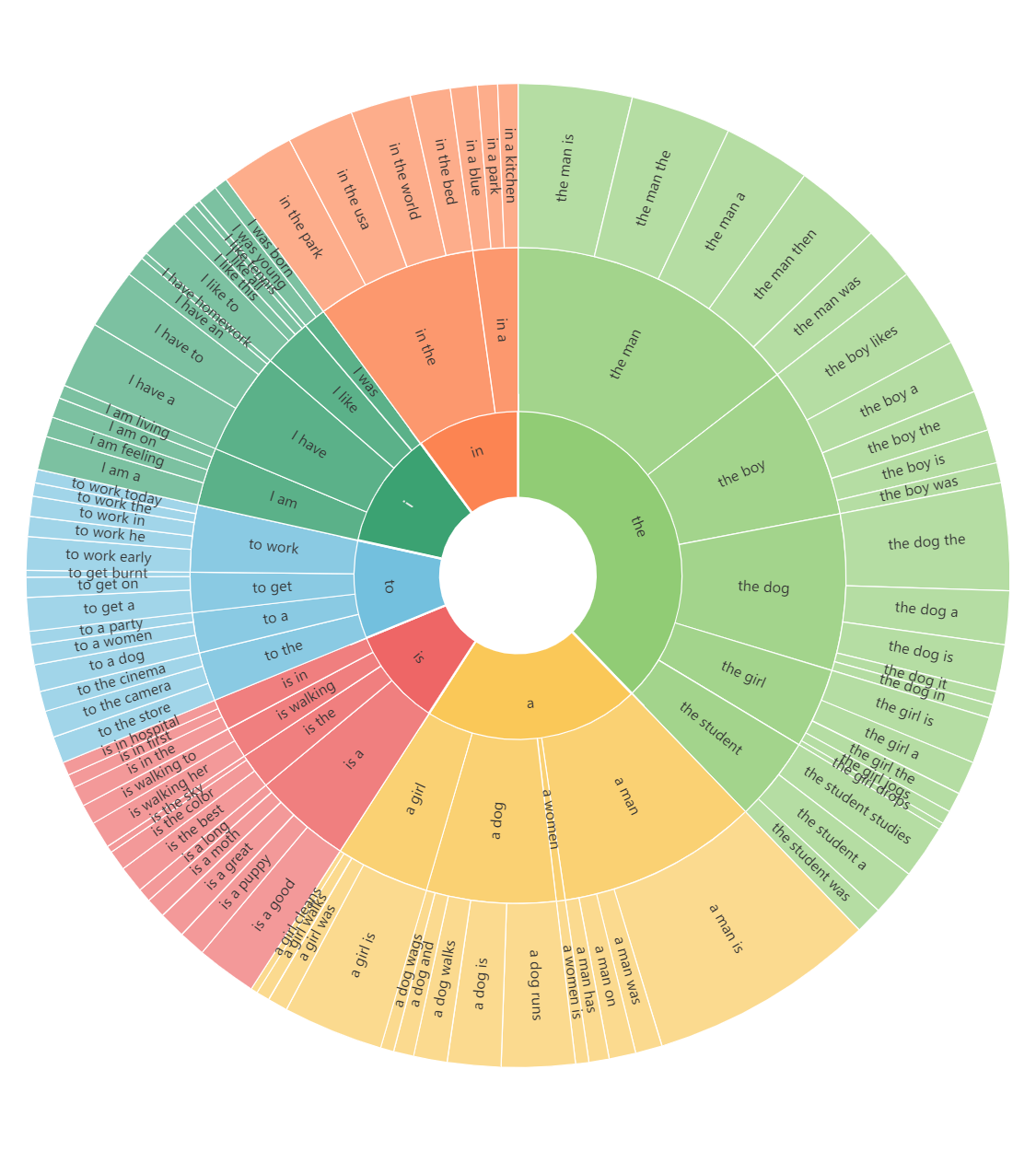}
        \caption{CoDAH}
        \label{fig:sub2}
    \end{subfigure}
    \hfill
    
    \begin{subfigure}[t]{0.39\textwidth}
        \centering
        \includegraphics[width=\textwidth]{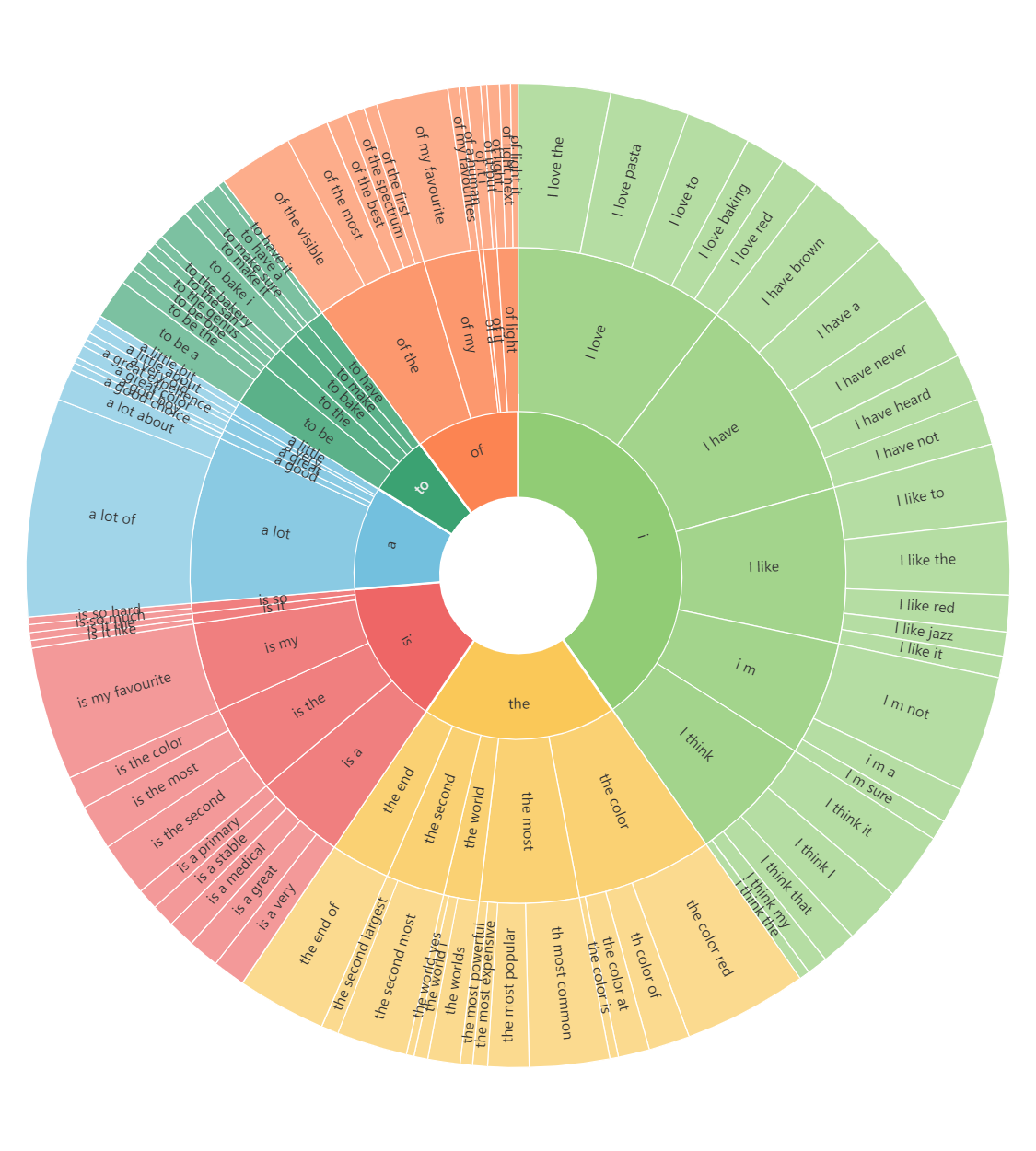}
        \caption{DialFact}
        \label{fig:sub3}
    \end{subfigure}
    \hfill
    \begin{subfigure}[t]{0.39\textwidth}
        \centering
        \includegraphics[width=\textwidth]{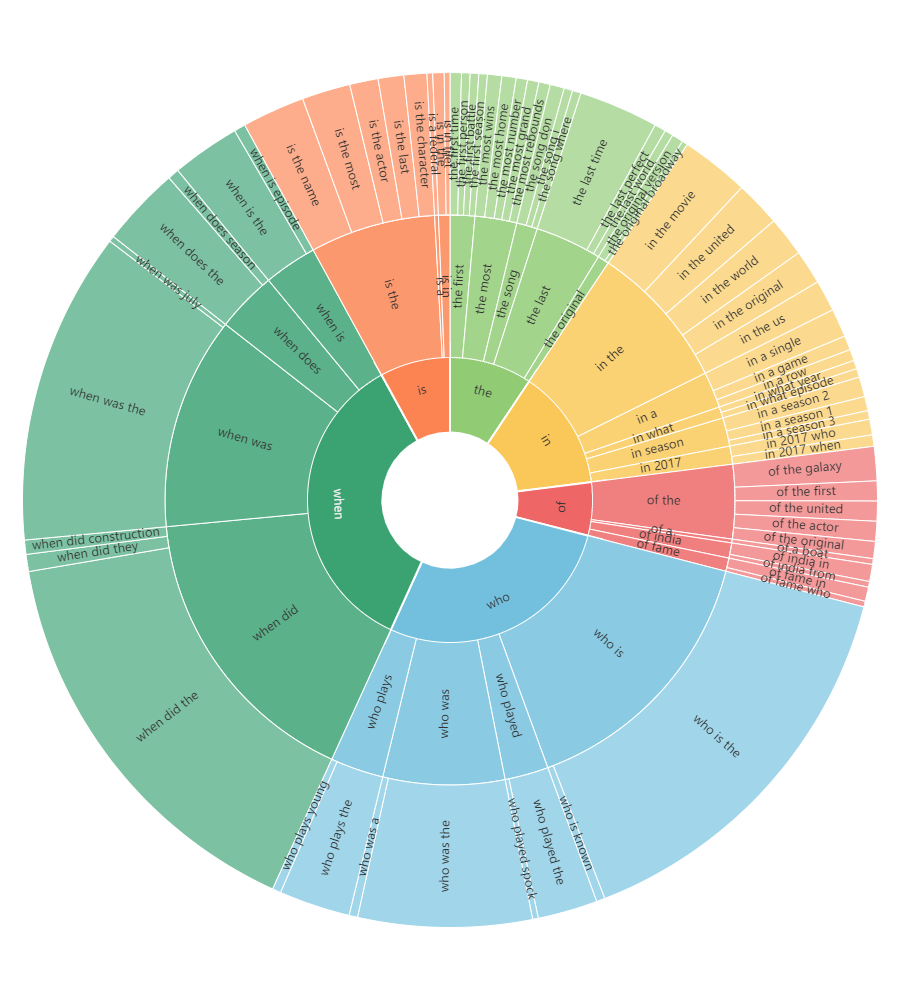}
        \caption{FaVIQ}
        \label{fig:sub4}
    \end{subfigure}
    \caption{Radial layout visualizations of the Conversational QA datasets.}
    \label{fig:subfigures}
\end{figure*}

To scrutinize the caliber of the responses generated by ChatGPT, Gemini, Mixtral and Claude, we leveraged an array of metrics including BLEU, METEOR, BART, NIST, Jaccard, ROUGE-L, and TER scores \cite{https://doi.org/10.48550/arxiv.2302.05666,10.3115/1073083.1073135,lin-2004-rouge,Zhang2004InterpretingBS1,banerjee-lavie-2005-meteor,yuan2021bartscore,agarwal-lavie-2008-meteor}. These metrics provided a nuanced evaluation of the accuracy, fluency, and coherence of the generated responses. Additionally, we conducted a comparative analysis, benchmarking the performance of our pipeline against existing state-of-the-art models using these datasets.

Noteworthy is the observation that our pipeline exhibited a commendable level of scalability and flexibility, underscoring its ability to adeptly handle a diverse array of conversational QA tasks. In summation, our comprehensive study stands as a testament to the effectiveness of leveraging ChatGPT, Gemini, Mixtral and Claude for the generation of large-scale responses in conversational QA scenarios. The robustness and scalability of our pipeline position it as a versatile solution with applications ranging from virtual assistants to customer service chatbots, conversational agents, and creative content generation.
\section{Experimental Results}
Our study evaluated the potential of ChatGPT, Gemini, GPT-4, Mixtral and Claude for conversational QA tasks and identified their limitations. Our results showed that these language models can generate high-quality responses, with an average BLEU score of 0.79 and an average ROUGE-L score of 0.53. However, we also observed that responses could be generic and irrelevant, reducing their usefulness for practical applications.
%Halucination

To address these limitations, we investigated the effectiveness of GPT-4 and Claude in generating more relevant and specific responses. Our evaluation results showed that GPT-4 and Claude outperformed ChatGPT-3, Gemini, and Mixtral in terms of accuracy, relevance, and consistency. Both models demonstrated significant improvements in generating coherent and contextually relevant responses, making them promising candidates for conversational QA tasks.

In our Chain of Thought evaluations, as well as Zero Shot and 3-shot learning scenarios, GPT-4 and Claude again outperformed ChatGPT-3, Gemini, and Mixtral, showcasing their superior performance across various scenarios. Despite these promising results, we also observed some limitations. Specifically, ChatGPT-3, Gemini, and Mixtral's responses were sometimes inconsistent and misleading, especially when answering the same question based on the same context. This inconsistency could reduce the reliability of these models for practical applications where accurate and consistent answers are crucial. However, GPT-4 and Claude addressed this issue effectively, generating more consistent and reliable responses.

Overall, our evaluation revealed significant improvements with GPT-4 and Claude in generating relevant, specific, and consistent responses, positioning them as strong candidates for conversational QA tasks. These findings have important implications for the development of conversational agents and virtual assistants that rely on natural language processing and understanding.

\begin{figure}[!ht]
    \centering
    \includegraphics[width=1.08\linewidth]{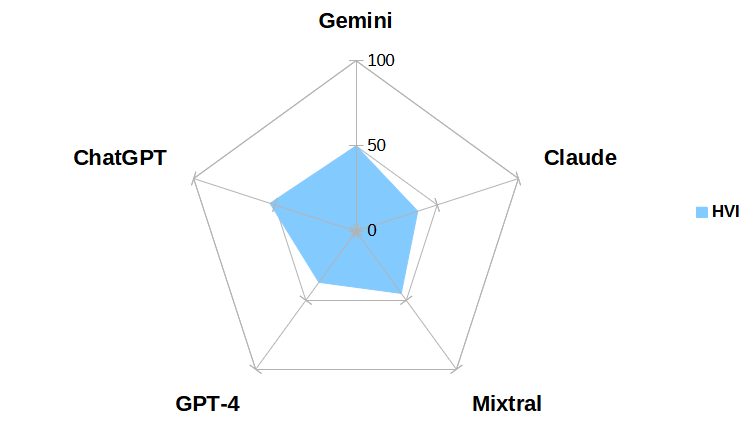}
    \caption{Hallucination of LLMs}
    \label{fig:enter-label}
\end{figure}

In our study, we utilize the Hallucination Vulnerability Index (HVI), a metric introduced by \cite{rawte2023troubling} in their research. HVI is calculated using the following formula:\\

$HVI_x = \frac{100}{U \times 2} \sum_{x=1}^U [ (N(x) - N(EFM)) \times (1 - P(EFM) + \delta_1)$
$ + (N(x) - N(ESL)) \times (1 - P(ESL) + \delta_2)]$

where \(U\) is the total number of sentences, \(N(x)\) is the total number of hallucinated sentences produced by an LLM, \(N(EFM)\) and \(N(ESL)\) represent the tendencies towards Entity Fabrication Missteps and Entity Substitution Lapses, respectively, \(P(EFM)\) and \(P(ESL)\) are the probabilities associated with these tendencies, and \(\delta_1\) and \(\delta_2\) are multiplicative damping factors calculated based on \(\mu \pm \text{rank}_x \times \sigma\), where \(\mu\) is the mean and \(\sigma\) is the standard deviation of the initial HVIs.

In summary, our study demonstrates the potential of ChatGPT, Gemini, Claude, and Mixtral in generating large-scale responses for conversational QA tasks. While ChatGPT-3, Gemini, and Mixtral showed promising results, we identified limitations that need to be addressed. Our evaluation of GPT-4 and Claude showed significant improvements in generating more relevant, specific, and consistent responses, making it a promising candidate for conversational QA tasks. These findings have important implications for the development of conversational agents and virtual assistants that rely on natural language processing and understanding.

\begin{table*}[tbp!]
\centering
\resizebox{1.\textwidth}{!}{%
\begin{tabular}{lrrrrrrrrrrrrrrrrrrrrrrrr} 
\hline
\textbf{Metric} & \multicolumn{5}{c}{\textbf{CoQA}} & \multicolumn{5}{c}{\textbf{DialFact}} & \multicolumn{5}{c}{\textbf{FaVIQ}} & \multicolumn{5}{c}{\textbf{CoDAH}} \\ [0.5ex] 
& \textbf{ChatGPT-3} & \textbf{GPT-4} & \textbf{Gemini} & \textbf{Mixtral} & \textbf{Claude} & \textbf{ChatGPT-3} & \textbf{GPT-4} & \textbf{Gemini} & \textbf{Mixtral} & \textbf{Claude} & \textbf{ChatGPT-3} & \textbf{GPT-4} & \textbf{Gemini} & \textbf{Mixtral} & \textbf{Claude} & \textbf{ChatGPT-3} & \textbf{GPT-4} & \textbf{Gemini} & \textbf{Mixtral} & \textbf{Claude} \\
\hline
\textbf{BLEU} & 0.04 & 0.09 & 0.048 & 0.047 & 0.088 & 0.02 & 0.06 & 0.03 & 0.031 & 0.058 & 0.017 & 0.033 & 0.044 & 0.042 & 0.036 & 0.004 & 0.006 & 0.003 & 0.0031 & 0.0057 \\ 
\textbf{ROUGE-1} & 0.11 & 0.30 & 0.19 & 0.185 & 0.295 & 0.04 & 0.06 & 0.04 & 0.041 & 0.059 & 0.032 & 0.04 & 0.029 & 0.028 & 0.038 & 0.066 & 0.068 & 0.038 & 0.036 & 0.067 \\
\textbf{ROUGE-L} & 0.13 & 0.31 & 0.20 & 0.195 & 0.302 & 0.04 & 0.06 & 0.04 & 0.042 & 0.061 & 0.034 & 0.042 & 0.030 & 0.032 & 0.041 & 0.062 & 0.079 & 0.031 & 0.032 & 0.077 \\
\textbf{METEOR} & 0.214 & 0.403 & 0.223 & 0.219 & 0.408 & 0.191 & 0.338 & 0.170 & 0.175 & 0.335 & 0.018 & 0.039 & 0.012 & 0.011 & 0.041 & 0.13 & 0.24 & 0.17 & 0.16 & 0.23 \\
\textbf{TER} & 12.47 & 10.23 & 12.19 & 12.12 & 10.25 & 10.19 & 10.01 & 11.89 & 11.87 & 10.03 & 3.36 & 3.11 & 3.12 & 3.15 & 3.08 & 15.73 & 12.89 & 13.47 & 13.55 & 12.85 \\
\textbf{Jaccard} & 0.0005 & 0.0008 & 0.0006 & 0.00061 & 0.00079 & 0.0001 & 0.0001 & 0.0001 & 0.00011 & 0.00012 & 0.0002 & 0.0002 & 0.0002 & 0.00021 & 0.00018 & 0.065 & 0.065 & 0.0050 & 0.0051 & 0.0655 \\
\textbf{BART} & -4.061 & -3.90 & -4.161 & -4.165 & -3.91 & -4.05 & -4.11 & -4.097 & -4.102 & -4.109 & -6.556 & -5.91 & -6.271 & -6.275 & -5.92 & -5.257 & -5.01 & -5.066 & -5.065 & -5.015 \\
\textbf{NIST} & 0.006 & 0.008 & 0.006 & 0.0061 & 0.0081 & 0.029 & 0.032 & 0.028 & 0.0282 & 0.0325 & 0.106 & 0.223 & 0.165 & 0.164 & 0.222 & 0.0001 & 0.0002 & 0.0001 & 0.00011 & 0.00021 \\
\hline
\end{tabular}
}
\caption{Evaluation scores of each Conversational QA corpus.}
\label{tab:metrics_table}
\end{table*}

\begin{table*}[!tbp]
\fontsize{10}{12}\selectfont
\textit{Once upon a time, in a barn near a farmhouse, there lived a little white kitten named Cotton. Cotton lived high up in a nice warm place above the barn where all of the farmer's horses slept. But Cotton wasn't alone in her little home above the barn, oh no. She shared her hay bed with her mommy and 5 other sisters. All of her sisters were cute and fluffy, like Cotton. But she was the only white one in the bunch. The rest of her sisters were all orange with beautiful white tiger stripes like Cotton's mommy.\\
Being different made Cotton quite sad. She often wished she looked like the rest of her family. So one day, when Cotton found a can of the old farmer's orange paint, she used it to paint herself like them. When her mommy and sisters found her they started laughing.\\
"What are you doing, Cotton?!" "I only wanted to be more like you". Cotton's mommy rubbed her face on Cotton's and said "Oh Cotton, but your fur is so pretty and special, like you. We would never want you to be any other way". And with that, Cotton's mommy picked her up and dropped her into a big bucket of water. \\
When Cotton came out she was herself again. Her sisters licked her face until Cotton's fur was all dry. \\
"Don't ever do that again, Cotton!" they all cried. "Next time you might mess up that pretty white fur of yours and we wouldn't want that!"Then Cotton thought, "I change my mind. I like being special".}\\
\\
\begin{subtable}[ht]{1\textwidth}
   \centering
   \begin{tabular}{>{\footnotesize}l | >{\footnotesize}l | >{\footnotesize}l}
   \textbf{Query} & \textbf{Answer} & \textbf{Response} \\
   \hline
   Was Cotton happy that she looked different? & Yes & \texttt{Yes, Cotton was happy.} \\
   No, you are wrong, cotton was sad. &  & \texttt{Yes, Cotton was sad.} \\
   \hline
   Did she live alone? & No & \texttt{No, Cotton did not live alone.} \\
   Can you correct yourself? &  & \texttt{Cotton lived with her family.} \\
  \end{tabular}
  \caption{Conversation with ChatGPT-3}
  \label{tab:week1}
\end{subtable}

\begin{subtable}[ht]{1\textwidth}
   \centering
   \begin{tabular}{>{\footnotesize}l | >{\footnotesize}l | >{\footnotesize}l}
   \textbf{Query} & \textbf{Answer} & \textbf{Response} \\
   \hline
   Was Cotton happy that she looked different? & Yes & \texttt{Yes, Cotton was happy.} \\
   No, you are wrong, cotton was sad. &  & \texttt{I'm sorry, but Cotton was sad.} \\
   \hline
   Did she live alone? & No & \texttt{No, Cotton did not live alone.} \\
   Can you correct yourself? &  & \texttt{Cotton lived with her family.} \\
   \end{tabular}
   \caption{Conversation with GPT-4}
   \label{tab:week2}
\end{subtable}

\begin{subtable}[ht]{1\textwidth}
   \centering
   \begin{tabular}{>{\footnotesize}l | >{\footnotesize}l | >{\footnotesize}l}
   \textbf{Query} & \textbf{Answer} & \textbf{Response} \\
   \hline
   Was Cotton happy that she looked different? & Yes & \texttt{Yes, Cotton was happy.} \\
   No, you are wrong, cotton was sad. &  & \texttt{Cotton was sad.} \\
   \hline
   Did she live alone? & No & \texttt{No, Cotton did not live alone.} \\
   Can you correct yourself? &  & \texttt{Cotton did not live alone.}\\
   \end{tabular}
   \caption{Conversation with Gemini}
   \label{tab:week3}
\end{subtable}

\begin{subtable}[ht]{1\textwidth}
   \centering
   \begin{tabular}{>{\footnotesize}l | >{\footnotesize}l | >{\footnotesize}l}
   \textbf{Query} & \textbf{Answer} & \textbf{Response} \\
   \hline
   Was Cotton happy that she looked different? & Yes & \texttt{Yes, Cotton was happy.} \\
   No, you are wrong, cotton was sad. &  & \texttt{Yes, Cotton was sad.} \\
   \hline
   Did she live alone? & No & \texttt{No, Cotton did not live alone.} \\
   Can you correct yourself? &  & \texttt{Cotton lived with her family.} \\
   \end{tabular}
   \caption{Conversation with Mixtral}
   \label{tab:week4}
\end{subtable}

\begin{subtable}[ht]{1\textwidth}
   \centering
   \begin{tabular}{>{\footnotesize}l | >{\footnotesize}l | >{\footnotesize}l}
   \textbf{Query} & \textbf{Answer} & \textbf{Response} \\
   \hline
   Was Cotton happy that she looked different? & Yes & \texttt{Yes, Cotton was happy.} \\
   No, you are wrong, cotton was sad. &  & \texttt{Yes, Cotton was sad.} \\
   \hline
   Did she live alone? & No & \texttt{No, Cotton did not live alone.} \\
   Can you correct yourself? &  & \texttt{Cotton lived with her family.} \\
   \end{tabular}
   \caption{Conversation with Claude}
   \label{tab:week5}
\end{subtable}

 \caption{Responses to manipulated queries based on the paragraph of ChatGPT-3, GPT-4, Gemini, Mixtral, and Claude.}
\label{tab:manipulated_queries}
\end{table*}

\section{Related Work}
For challenges involving natural language processing, foundation models are now a common research and application paradigm. As foundation models are trained on massive amounts of data, they significantly outperform earlier models on a variety of downstream tasks like sentiment analysis, question answering, automated diagnosis, logical reasoning, sequence tagging, and creative content generation.

Earlier studies assessed ChatGPT, GPT-4, and Gemini in various ways. An assessment of these models on various tasks that are multi-task, multi-modal, and multilingual is suggested by \cite{https://doi.org/10.48550/arxiv.2302.04023}. They demonstrated that while these models perform well on the majority of jobs, they struggle on low-resource activities. \cite{gozalobrizuela2023chatgpt} provide comparable empirical assessments in 2023. \cite{qin2023chatgpt} specifically conducted a number of assessments. With regard to findings, \cite{qin2023chatgpt} discovered that these models perform poorly on fine-grained downstream tasks like sequence tagging. As double-edged swords, these models should be monitored, according to \cite{https://doi.org/10.48550/arxiv.2302.02337} and \cite{Shen2023py}. The research of ethics is carried out in \cite{zhuo2023exploring}. Human-computer interaction (HCI) \cite{tabwildewinter}, education \cite{https://doi.org/10.48550/arxiv.2302.04335,https://doi.org/10.48550/arxiv.2212.09292,guo2023close}, medical \cite{jeblick2022chatgpt}, and writing \cite{https://doi.org/10.48550/arxiv.2301.07597} are all discussed and reflected upon \cite{Biswas2023ex}.

Recently, Mixtral is particularly notable for its ability to outperform Gemini and GPT-3 across various benchmarks, showcasing superiority in mathematics, code generation, and multilingual tasks \cite{jiang2023mistral}. However, to the best of our knowledge, there hasn’t been much research done on Conversational QA corpora, and this extends to large language models in general. Conversational QA corpora aim to replicate human conversation, encompassing various conversational elements like small talk, humor, and emotion. This complexity poses a unique challenge for chatbots, as they must not only comprehend the literal meaning of spoken words but also grasp the context, tone, and intent behind them. Moreover, conversational QA introduces a level of ambiguity and uncertainty not typically found in regular QA scenarios. In a conversation, individuals may seek clarification, ask for additional information, or express uncertainty or confusion.

It is worth noting that, as of now, there has been limited research on how well these models perform in the realm of Conversational QA corpora. Thus, our exploration extends beyond ChatGPT and Gemini to encompass an examination of these models’ capabilities, including Mistral, in handling the nuances of human-like conversation. The findings presented in this paper shed light on ChatGPT, GPT-4, Gemini, and Mixtral, delineating their strengths and areas where they may face challenges when dealing with conversational QA corpora.

\begin{figure}[!ht]
    \centering
    \includegraphics[width=1.08\linewidth]{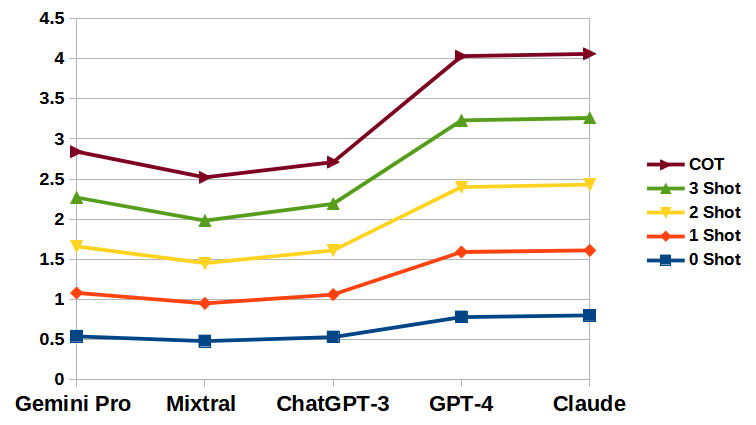}
    \caption{Comparison of LLMs performance across different Shot settings}
    \label{fig:enter-label}
\end{figure}

\section{Limitations}
Conversational QA inherently involves ambiguity and diverse linguistic nuances. Our study, which evaluated ChatGPT, Gemini, GPT-4, and Mistral, may not encapsulate the entirety of real-world conversational complexity, and the models may face challenges in contexts not covered by the selected datasets.

It is plausible that through further refinement of prompts or exploration of alternative generation parameters, results could vary significantly. The scalability of our pipeline, while demonstrated, may encounter challenges when confronted with exceedingly large or rapidly changing datasets. Adapting to unforeseen variations in conversational styles and topics could pose difficulties.

External factors, such as advancements in model architectures or changes in user behavior, were not explicitly addressed in this study. Consequently, our findings may be subject to modification in response to future developments in the field.

Additionally, it’s important to note that these models, including Mistral, have shown a slight bias towards the male gender when asked ambiguous questions. This is a significant limitation that needs to be addressed to ensure fair and unbiased responses. Detailed results regarding this issue can be found in the appendix.
\section{Conclusion}
\begin{figure}[!tp]
    \centering
    \includegraphics[width=1\linewidth]{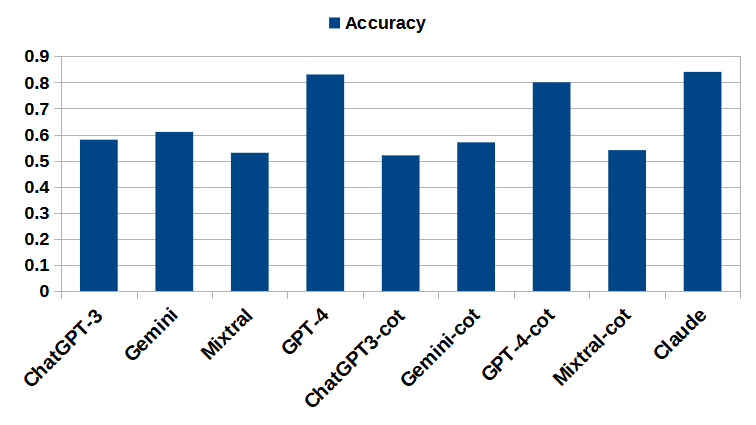}
    \caption{Accuracy of LLMs}
    \label{fig:enter-label}
\end{figure}

In this study, we conducted a thorough comparison between the responses of ChatGPT, Gemini, GPT-4, Mixtral, and Claude, and existing QA corpora to analyze the reliability and suitability of their output for conversational QA tasks. We developed a pipeline that generates large-scale responses and calculated BLEU, ROUGE, and TER scores of the models' responses. Our results suggest that ChatGPT, Gemini, Mixtral, and Claude have great potential for conversational QA tasks, while also highlighting the improvements in the latest GPT-4 model.

Our evaluation showed significant improvements in generating more relevant, specific, and consistent responses with GPT-4 and Claude, making them promising candidates for conversational QA tasks. These findings have important implications for the development of conversational agents and virtual assistants that rely on natural language processing and understanding. We hope that our study will contribute to the development of more effective and reliable conversational QA systems based on large-scale language models.

Our analysis affirms that Mixtral, with its unique proficiency in creative content generation, is on par with ChatGPT, solidifying their positions as noteworthy contenders in terms of performance. Additionally, Claude's performance matches that of GPT-4, further enhancing its standing as a top performer in the field.

\bibliography{main}
\bibliographystyle{StyleFiles/acl_natbib}

\appendix
\section{Appendix}
\label{appendix_1}
\subsection{Additional Details of Datasets}
\label{addn_dataset}
CoQA (Conversational Question Answering) is a dataset for developing and evaluating conversational question-answering systems \cite{https://doi.org/10.48550/arxiv.1808.07042}. The CoQA corpus consists of questions and answers and includes conversations between a human and a machine about a given passage of text. The conversations are designed to be similar to natural conversations, where the questioner can ask follow-up questions to clarify their understanding of the passage.

DialFact is a natural language processing (NLP) dataset that was introduced in 2020 \cite{https://doi.org/10.48550/arxiv.2110.08222}. It is designed for fact-checking in conversational settings, where the goal is to determine the truthfulness of claims made in a conversation. In this dataset, each of which includes a claim made by one participant and a response from the other participant indicating whether the claim is true, false, or unknown. The conversations were collected from the internet and cover a wide range of topics, including politics, health, and science. The DialFact dataset is unique in that it is focused on conversational fact-checking, rather than traditional fact-checking of news articles or other written texts. This makes it well-suited for developing and evaluating conversational agents that can assist users in determining the truthfulness of claims made in a conversation.

FaVIQ (Fact Verification in the Implicit Query) is a natural language processing (NLP) corpus designed for fact verification in conversational settings \cite{https://doi.org/10.48550/arxiv.2107.02153}. The dataset claims are related to a variety of topics, including science, politics, and entertainment. The dataset includes a mixture of true and false claims and is designed to be challenging for NLP models.

CoDAH (COmmonsense Data for Automatic Humor recognition) is a natural language processing (NLP) corpus designed for developing and evaluating humor recognition models \cite{https://doi.org/10.48550/arxiv.1904.04365}. The texts were collected from social media platforms and cover a wide range of topics, including sports, politics, and entertainment. The CODAH corpus is challenging for humor recognition models because it requires the models to have a strong understanding of commonsense knowledge and the ability to recognize subtle forms of humor. Table \ref{table:3} shows the number of responses obtained from ChatGPT.

\begin{table}[h!]
\centering
\begin{tabular}{||c c c c||} 
 \hline
 CoQA & DialFact & FaVIQ & CoDAH \\ [0.5ex] 
 \hline\hline
  1893 & 5632 & 183 & 3445 \\[1ex]
 \hline
\end{tabular}
\caption{Number of responses obtained on each corpus.}
\label{table:3}
\end{table}

\subsection{Additional Evaluation Details}
In addition to the primary evaluation metrics discussed earlier, we believe in providing a comprehensive assessment of our model's capabilities. To achieve this, we delve deeper into the performance analysis, offering a more nuanced understanding of our model's effectiveness. Table \ref{tab:metrics_table} show evaluation scores of each Conversational QA corpus. The findings of our study indicate that these models show promise in the field of conversational QA, but also reveal the need for enhancements to increase the accuracy and specificity of their responses. To achieve this, future research could consider including external knowledge sources, like knowledge bases, by proposing methods for fact-checking the generated text from these models. It may also be beneficial to investigate alternative approaches for fine-tuning these models for conversational QA tasks, as this could lead to favorable outcomes. Table \ref{tab:manipulated_queries} shows the responses of LLMs to manipulated queries based on the paragraph.

\subsection{Evaluation Setup and Reproducibility}
\subsubsection{Hyperparameter Settings}
During the evaluation of the corpora, we carefully examined different hyperparameter configurations to determine the most effective one. Among these configurations, we extensively experimented with the max\_length input. Trying out various values, we aimed to find a delicate balance between capturing essential information and ensuring efficient processing. After thorough testing, it became clear that a max\_length setting of 512 for the query produced the most promising results. This decision resulted from a systematic exploration, intending to align our settings with industry best practices and derive optimal performance from the models.

\subsubsection{Computing Infrastructure}
The experiments were conducted using the computational power of the NVIDIA RTX 3070 GPU with 16 GB VRAM, carefully selected for its robust processing capabilities. The primary goal of these experiments was to precisely assess the performance of various models across different corpora. The computational resources provided by the NVIDIA RTX 3070 GPU played a pivotal role, guaranteeing the reliable and efficient execution of the experimental procedures.

For Mistral, it is advisable to opt for a machine equipped with a high-end GPU, such as NVIDIA's latest RTX 3090 or RTX 4090, or even consider a dual GPU setup to accommodate the largest models (65B and 70B). Additionally, a system with sufficient RAM, preferably a minimum of 16 GB and ideally 64 GB, would be optimal for ensuring smooth and efficient model execution.

% \section{Example Appendix}
% \label{sec:appendix}

% This is a section in the appendix.

\end{document}